%% file: main.tex
\documentclass[11pt]{article}

\usepackage[margin=1in]{geometry}

\usepackage{times}
\usepackage{graphicx}
\usepackage{float}
\usepackage{placeins}
\usepackage{authblk}
\usepackage{hyperref}
\usepackage{orcidlink}
\usepackage{amsmath}

\title{\bf CALM: Contextual Analog Logic with Multimodality}

\author[1]{Maxwell J. Jacobson\,\orcidlink{0000-0002-0888-028X}}
\author[1]{Corey J. Maley\,\orcidlink{0000-0001-6221-3181}}
\author[1]{Yexiang Xue\,\orcidlink{0000-0002-4533-0543}}
\affil[1]{Purdue University, \texttt{\{jacobs57, cjmaley, yexiang\}@purdue.edu}}

\date{}

\newcommand{\assignment}{ grounding }
\newcommand{\assignments}{ groundings }

\begin{document}

\maketitle

\begin{abstract}
In this work, we introduce Contextual Analog Logic with Multimodality (CALM). 
CALM unites symbolic reasoning with neural generation, enabling systems to make context-sensitive decisions grounded in real-world multi-modal data.

\textbf{Background:} 
Classic bivalent logic systems cannot capture the delicate human decision-making process.
They also require human grounding (interpretation) in multi-modal environments, which can be ad-hoc, rigid, and brittle.
Neural networks are good at extracting rich contextual information from multi-modal data, but lack interpretable structures for reasoning. 
Combining logic and neural networks at a deep, semantic level is required. 

\textbf{Objectives:}
CALM aims to bridge the gap between logic and neural perception, creating an analog logic that can reason over multi-modal inputs. 
Without this integration, AI systems remain either brittle or unstructured, unable to generalize robustly to real-world tasks.
In CALM, symbolic predicates evaluate to analog truth values computed by neural networks and constrained search. 

\textbf{Methods:}
CALM represents each predicate using a domain tree, which iteratively refines its analog truth value when the contextual groundings of its entities are determined.
The iterative refinement is predicted by neural networks capable of capturing multi-modal information and is filtered through a symbolic reasoning module to ensure constraint satisfaction. 

\textbf{Results:}
In fill-in-the-blank object placement tasks, CALM achieved 92.2\% accuracy with half of the spatial relationships left ambiguous, outperforming both a classical logic baseline (86.3\%) and an LLM baseline (59.4\%). It also demonstrated spatial heatmap generation aligned with logical constraints and delicate human preferences, as shown by a human study.
    
\textbf{Conclusions:}
CALM demonstrates the potential to reason with rigorous logic structure while aligning with subtle preferences in multi-modal environments. 
It lays the foundation for next-generation AI systems that require the precision and interpretation of logic and the delicate, multimodal information processing of neural networks.
\end{abstract}

\section{Introduction}

\input{tex/introduction}

\section{Background}
\input{tex/background}

\section{Contextual Analog Logic with Multimodality (CALM)}

\input{tex/method}

\section{Experiments}

\input{tex/experiments}

\section{Conclusion}

\input{tex/conclusion}

\section*{Acknowledgements}
This research was supported by NSF Career Award IIS-2339844, DOE – Fusion Energy Science grant: DE-SC0024583.

\FloatBarrier

\bibliographystyle{plain}
\bibliography{sample-base}

\end{document}

%% file: tex/introduction.tex
\begin{figure}
    \centering
    \includegraphics[width=0.95\linewidth]{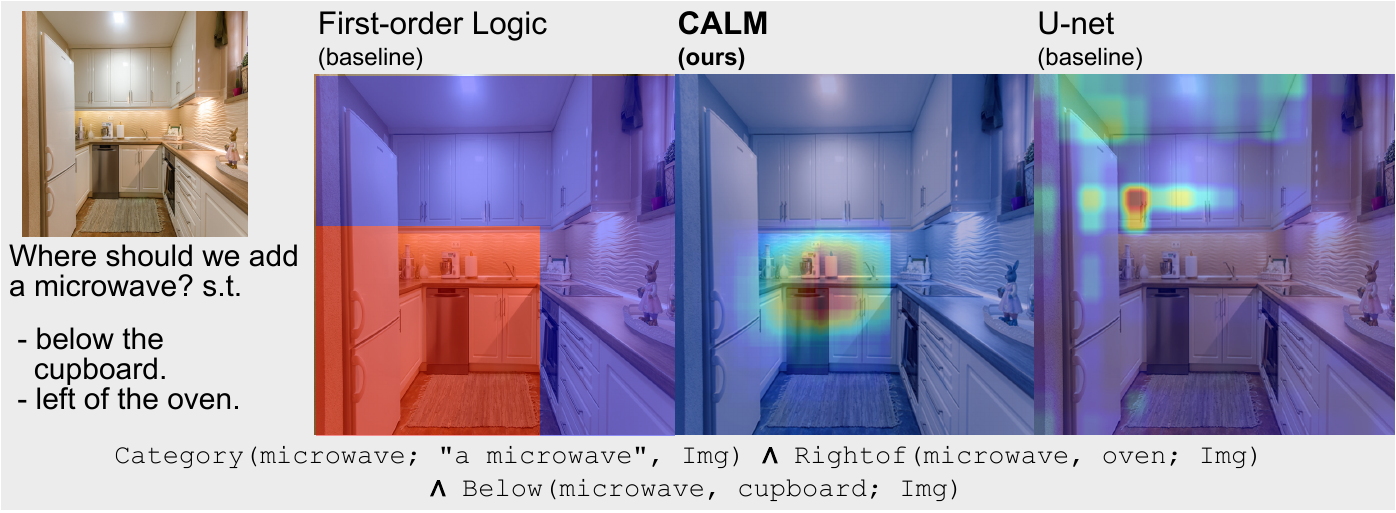}
    \caption{We introduce Contextual Analog Logic with Multimodality (CALM), which automatically grounds logic within multi-modal contexts. Here, the task is to decide where to add a microwave to a kitchen image subject to spatial constraints. First-order Logic (FOL) (left panel) outputs a correct solution, but it is not grounded in the surrounding context of the image, nor is it able to express preference based on that context due to bivalent truth values. A U-net \cite{unet} neural network (right panel) can learn to ground itself in multi-modal logic, but it does not guarantee a correct solution to the user's specification -- here suggesting locations above the cupboard. Our CALM outputs a correct solution with analog truth values, also grounded in the context of the image (area in warmer colors on the countertop in the middle panel). }
    \label{fig:fig1}
\end{figure}

Formal logic is a cornerstone of artificial intelligence (AI). 
With clearly defined syntactic forms, semantic meaning, and inference rules, logic provides robustness, composability, and interoperability, allowing us to derive useful consequences from given premises. 
As a result, logic is central to a wide array of AI applications  \cite{Prestwich2021,10.1007/978-3-540-30201-8_12}.

However, classical logic faces significant limitations in both expressiveness and grounding. 
%
%\xyx{Limitations of bivalent logic...} 
Classical bivalent logic, which assigns true-or-false values to predicates, is limited in capturing the delicate nature of many real-world reasoning tasks. 
%
%, has been studied since Aristotle and undeniably possesses utility, yet it cannot adequately represent many cognitive phenomena or real-world scenarios. 
Numerous predicates naturally require evaluation as degrees of truth rather than as binary values. Consider, for example, instructing a person to place a microwave ``below the cupboard'' and ``left of the oven,'' as shown in Figure~\ref{fig:fig1}. Implicit, context-dependent constraints affect the natural interpretation of these instructions. A continuous range of acceptable placements exists: the position ``below the cupboard'' is intuitively truer when the microwave rests directly on the countertop, rather than further down or closer to the floor. Classic logic would place the microwave in a region rigidly defined by the constraints, as shown in the red area in the left panel. Many locations within this range are not reasonable. On the contrary, as illustrated by the heat map in the middle panel of the figure, modeling such predicates with graded, real-number truth values rather than strict binaries captures this nuance more accurately. 

%\xyx{limitations with grounding...} 
Moreover, classic logic requires \textit{grounding}, or interpretation, in real-world scenarios. 
Currently, such grounding can only be performed by human beings. This seriously affects the applicability of automated reasoning. 
Grounding interprets the meaning of symbolic entities in real-world contexts. In our example, determining appropriate microwave placement depends on perceptual context, such as visually recognizing the spatial arrangements of objects in the kitchen. Without mechanisms to integrate real-world perception, logic remains abstract, and disconnected from contextually rich reasoning tasks.

Fuzzy logic \cite{richardson2006markov,zadeh_fuzzy_logic, 8543645} and probabilistic programming languages \cite{skryagin2023scalable,MANHAEVE2021103504,olausson-etal-2023-linc,pmlr-v70-bosnjak17a} have addressed the limitations in expressiveness by generalizing classical logic to handle degrees of truth, allowing predicates to take real values rather than strict binaries. Despite these improvements, fuzzy and probabilistic logics still do not ground symbolic reasoning within the unstructured data of the real world. 
In parallel, neural perception \cite{yolo,detr,kirillov2023segany} has recently achieved remarkable progress in extracting rich contextual information from multi-modal data such as images, text, and video. These neural approaches excel at predictive and generative tasks precisely because they learn latent representations directly from perceptual inputs. %, effectively linking symbolic concepts to the tangible contexts in which they occur. 
However, the learned representations often  lack interpretability, and hence they cannot fit together with a clearly-defined formal logic.

A critical gap remains: there is currently no effective method to integrate the multi-modal context-awareness of neural perception with the deductive reasoning ability of formal logic. 
As a result, symbolic reasoning systems remain artificially separated from perceptual input, struggling to generalize effectively in noisy, real-world scenarios that require semantic grounding or the resolution of ambiguity. Humans are forced to bridge this gap manually, explicitly defining symbolic representations and their grounding in rigid ways -- an approach that is brittle, unscalable, and unadaptive without extensive ad-hoc engineering.

We introduce \textbf{Contextual Analog Logic with Multimodality (CALM)}.
First, CALM is analog. Its predicates evaluate to continuous, analog truth values, allowing CALM to capture the delicacy of fine comparisons and preferences. See the middle panel of Figure \ref{fig:fig1}, where the locations of more probable microwave placements have higher truth values in CALM (in warmer colors).
Because the continuous truth values may change across different contexts, CALM is analog, rather than simply fuzzy. 
Second, CALM grounds logic entities in the multi-modal environment -- real-world data in the form of text, images, or other means of perception. 
Every symbol in CALM is associated with 
a multi-modal context. In our image generation application, such multi-modal contextual grounding is the bounding box showing the location of the referred object (a microwave, in the example in Fig. \ref{fig:fig1}). CALM achieves multi-modal grounding using a neural network to predict the analog truth value of a predicate when all of its entities are grounded in the multi-modal environment.

%
%Semantic grounding. 
%
%CALM achieves semantic grounding by allowing neural networks trained on real-world perceptual data to inform predicate truth values contextually. 

At a high level, the neural and symbolic reasoning are combined in the following way in CALM. 
%
%At a high level, the symbols in CALM are grounded in the multi-modal context. 
%
For every predicate, we define a domain tree that iteratively refines its truth value when the contextual \assignments of its entities are determined. 
For example, a predicate in the example in Fig. \ref{fig:fig1} is ``below(microwave, cupboard, Image)''. The root of the domain tree for this predicate places no restrictions on the placement of the microwave. The root may have two branches, one corresponding to placing the microwave above the cupboard and one below. 
Obviously, following the first branch incurs a big decrease in the truth value of this predicate, while following the second should incur an increase. 
CALM leverages a neural network in the prediction of such changes in the analog truth values. 
%
%\xyx{We also implement a constraint reasoning layer in the neural network, which guarantees that locations that clearly violate constraints receive a truth value of 0 (in this case, following the first branch)}. 
It also utilizes a structured constraint reasoning search, which guarantees that locations that clearly violate constraints receive a truth value of 0 (in this case, following the first branch).
Notice that even though every placement satisfies the hard constraint of placing the microwave underneath the cupboard after following the second branch, the neural net continues to adjust the truth value of the predicate while descending subsequent branches.
This allows CALM to capture the delicacy of logic reasoning in multi-modal environments. 
For example, it should be less likely that a microwave is placed close to the floor.
Such capability is beyond what traditional logic reasoning at an abstract level can offer.

%\xyx{Inference of CALM}
CALM supports three logic inference procedures. The first is truth evaluation: Given the \assignment of all entities, truth evaluation computes the truth value of the configuration. 
The second is truth maximization: given a multi-modal context (e.g., an image), this inference computes the \assignment of all objects which maximizes the truth value of a pre-defined CALM logic formula.
The third is truth-proportional sampling: given a multi-modal context, this inference samples the \assignment of all entities proportional to their truth values. 
%
%All inference procedures of CALM are accomplished through an importance sampling-fused backtrack search on the domain trees, similar in concept to the Sample-search \cite{samplesearch} \xyx{is this true? cite}. 
All inference procedures of CALM are accomplished through a variety of importance sampling-fused backtrack search methods on the domain trees.

%\xyx{Learning of CALM}
The learning of CALM is achieved by training the neural network to place objects in reasonable locations in given scenes.
While the objective meaning of many predicates is clear, learning is needed to recover the hidden, subtle aspect of many predicates. For example, ``below(microwave, cupboard, image)'' not only put the microwave strictly below the cupboard, but implicitly suggesting a location not too close to the floor. 
The meaning of a predicate can change with the participating items. For example, ``below(dog, cupboard, image)'' now should refer to a location on the floor as the object changes.
During training, we intentionally remove objects from existing images using inpainting, and ask the neural network to add them back. The neural net should be trained to assign high truth values to the original locations of these objects in the images.

%CALM is evaluated through a series of spatial reasoning tasks that test its ability to interpret and apply grounded logic. In a fill-in-the-blank setting, CALM assigns objects to occluded regions using both visual context and analog logic, outperforming both a vision-language model and a symbolic logic baseline -- especially under partial information. In this setting, a system controlled by CALM logic achieved an accuracy of 92.2\% at selecting the right blank even when only partial spatial logic was given. This surpasses the performance of a FOL-based system and an LLM-based system by 6 and 33 percentage points, respectively. In heatmap prediction, CALM generates interpretable spatial distributions that align with logical statements, unlike a U-net baseline which fails in correctness much of the time. Finally, in image inpainting, CALM samples object placements proportional to truth, enabling clean integration with generative models like Stable Diffusion \cite{rombach2021highresolution}. \xyx{add new experimental results when available.}

CALM is evaluated through a series of spatial reasoning tasks that test its ability to interpret and apply grounded logic. First, in a fill-in-the-blank task, CALM places missing objects into scenes using both visual context and logical constraints. It achieves 92.2\% accuracy even when only half the spatial relationships are provided—outperforming a classical logic system by 6 percentage points and a large vision-language model by 33. Second, CALM generates spatial heatmaps that show where an object should be placed to best match a given instruction. In a human study, participants rated CALM’s heatmaps as significantly more aligned with the provided logic than those of a neural network baseline, with strong statistical significance ($p < 0.0001$). Finally, CALM supports logic-guided image editing by sampling object placements proportional to their truth value under a given logical statement. These sampled placements are used to guide a generative model, such as Stable Diffusion \cite{rombach2021highresolution}, which then adds new objects into the image in a realistic and diverse way, while still respecting the specified spatial constraints.

Thus, CALM directly addresses the critical gap by imbuing logic with multi-modal context, extending classic logic in producing logic-based reasoning fit for real-world tasks without manual and brittle human-defined grounding. This work applies CALM in an interior design task where new objects must be placed within an image according to a logical statement and context about the image and the objects to be added.
However, CALM can be extended to other domains, for example, which robotic manipulation trajectories are safer and more natural, and hence should receive higher truth values while completing the task. 
This paper highlights CALM's usage in a visual domain. Such extensions will be left as future work.

\noindent This work's main contributions are:
\begin{itemize}
    \item \textbf{A new framework:} We define \textit{Contextual Analog Logic with Multimodality (CALM)}, a logic system that addresses the limited expressiveness and lack of grounding in classical logic, as well as the lack of structure and composability in neural models. CALM evaluates predicates to analog truth values grounded in multi-modal context, enabling interpretable and structured reasoning.

    \item \textbf{Three types of inference:} CALM supports efficient \textit{truth evaluation}, \textit{truth maximization}, and \textit{truth-proportional sampling}, enabling logic-based scoring, optimization, and probabilistic generation in perceptual contexts.
    
    \item \textbf{Empirical validation:} We found CALM excels in three settings: a fill-in-the-blank object placement task (92.2\% accuracy at 50\% logic, 6\% better than classic logic, and 33\% better than LLMs), a human-rated heatmap alignment study (better than baselines in logic satisfaction, with high significance $p < 0.0001$), and logic-constrained image inpainting.
\end{itemize}

%% file: tex/background.tex
\subsection{Multi-modal Neural Systems}

The recent era of deep learning has seen the rise of multi-modal systems -- models capable of processing and integrating information from diverse unstructured data types such as images, text, video, and 3D structures. These systems reflect a fundamental shift in AI: from single-modality pattern recognition toward unified representation learning across perceptual domains. This shift has made it possible for models to perform tasks involving rich, overlapping context, such as aligning textual instructions with visual content or grounding linguistic concepts in physical space.

Early progress in this area relied on specialized architectures. Autoencoders and convolutional neural networks (CNNs) provided the foundation for extracting compact and meaningful features from visual data \cite{convnets1,convnets2}. Later, transformers \cite{vaswani2017attention} extended this approach through sequence modeling via self-attention, allowing models to track long-range dependencies. The transformer architecture, when scaled, enabled large language models (LLMs) to capture abstract patterns across text and formed the basis for unifying multi-modal inputs.

One of the most influential steps toward true multi-modal integration was CLIP \cite{radford2021learning}, which learned to embed images and natural language into a shared vector space. This latent alignment allows semantically similar concepts -- like a photo of a cat and the phrase ``a small feline'' -- to occupy nearby regions in the embedding vector space. CLIP supports zero-shot classification and retrieval, but its broader significance lies in providing general-purpose, pre-trained encoders that make multi-modal grounding accessible to downstream systems. 

\subsection{Formal Logic}

Formal logic provides a systematic foundation for reasoning by defining explicit symbols, syntax, and inference rules that guarantee truth preservation from premises to conclusions. At its simplest level, propositional logic treats entire statements as indivisible units, connecting them with logical operators such as conjunction ($\land$) or disjunction ($\lor$). First-order logic (FOL) extends propositional logic significantly, introducing variables, predicates, and quantifiers, allowing generalized claims about objects. For example, "All dogs are mammals" expressed formally as $\forall x, \text{Dog}(x) \rightarrow \text{Mammal}(x)$. These logical systems, are rooted in classical bivalent logic. Every statement is evaluated strictly as true or false based on its interpretation within a given model.

\subsection{Fuzzy Logic}

Fuzzy logic extends classical logic by allowing truth values to range continuously between 0 and 1, enabling reasoning with vague or imprecise concepts \cite{richardson2006markov,zadeh_fuzzy_logic,8543645}. It has been successfully applied in areas such as control systems, process automation, and robotics, where approximate reasoning is often more effective than hard logic \cite{PRECUP2011213}. Over decades, fuzzy logic has formed into several distinct families. Zadeh’s original fuzzy logic began with fuzzy set theory \cite{ZADEH1965338} and introduced fuzzy constraints as a way to model imprecision formally \cite{doi:10.1287/mnsc.17.4.B141,zadeh1975calculus}. Type-2 fuzzy logic extended this by allowing the membership functions themselves to be fuzzy, capturing second-order uncertainty about the degree of truth \cite{zadeh1975linguistic1,zadeh1975linguistic2,zadeh1975linguistic3}. Mamdani fuzzy logic defined a rule-based system with fuzzy outputs, which are aggregated and then defuzzified to produce crisp control actions \cite{MAMDANI19751}. This architecture remains popular in industrial applications due to its human interpretability. TSK fuzzy logic replaced fuzzy outputs with crisp mathematical functions, often linear, making it more suitable for integration with learning systems \cite{takagi1985fuzzy}. ANFIS \cite{jang1993anfis} built on this by using early neural networks to learn membership functions and rule parameters directly from data, representing an early neuro-fuzzy system.

In more recent work, hybrid neuro-fuzzy systems have extended these ideas by using neural networks to learn fuzzy membership functions and rules from data, enhancing both robustness and interpretability \cite{talpur2023deep}. Some methods achieve highly accurate TSK-based classifiers that retain transparency even on large datasets \cite{7555341}. A natural next step for this line of work is to explore integration with symbolic programming and logical composition, particularly in the context of multimodal systems.

\subsection{Neuro-symbolic Probabilistic Programming Languages (NS-PPL)}

In a similar direction, neuro-symbolic probabilistic programming languages, aim to unify learning and reasoning by embedding neural networks into structured programming languages. In some such systems \cite{MANHAEVE2021103504}, predicates are defined by neural classifiers that return probability values based on fixed local input. These outputs are then composed with traditional logical inference, enabling the system to learn from data while maintaining symbolic structure. Other approaches in this space focus on translating natural language into logic or combining partial programs with learned components. In these systems, LLMs generate multiple interpretations of a query, which are then resolved via logical inference and voting \cite{olausson-etal-2023-linc}, or programmers may leave blanks in program structure to be filled in by neural modules trained from examples \cite{pmlr-v70-bosnjak17a}. These designs exemplify the broader goal of unifying symbolic and neural reasoning, and they represent important steps toward that vision. While many current systems operate with limited contextual grounding, the field is growing, and such capabilities will likely become increasingly central as the research landscape develops.

%% file: tex/method.tex
The goal of CALM is to enable logic-based reasoning that is semantically grounded in real-world perceptual contexts, bridging the advantages of logical reasoning and multi-modal neural networks. Recall the example given in the introduction. CALM allows us to logically express conditions like placing a microwave ``below the cupboard and left of the oven'' while simultaneously evaluating these statements within the real-world visual context of the kitchen --  taking into account elements like the countertop, the floor, and the semantic implications of these visual signals. This is achieved by integrating multi-modal neural networks directly into analog logic predicate evaluation.  

\begin{figure}[htbp]
    \centering
    \includegraphics[width=0.75\linewidth]{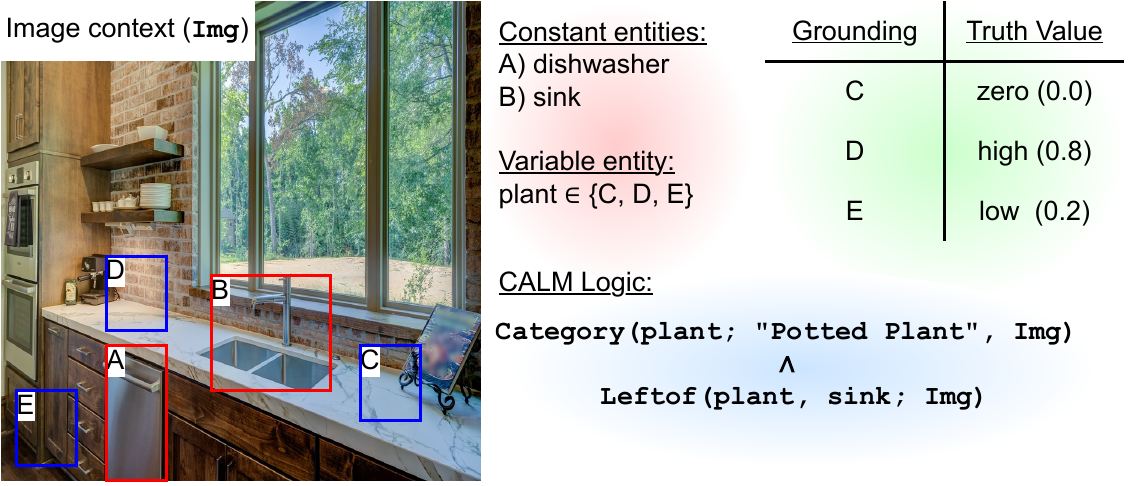}
    \caption{An example problem CALM can solve: Given a kitchen scene and a natural language description, CALM determines where to place a new object so that it satisfies both categorical and spatial constraints. In this case, the goal is to place a potted plant somewhere in the scene such that it matches the description "Potted Plant" and is to the left of the sink. The logic statement combines two predicates (\texttt{Category} and \texttt{Leftof}) with a conjunction connective ($\land$). The entities involved include two constant entities -- the dishwasher (A) and the sink (B) -- and one variable entity, plant, which can be assigned one of three candidate locations (C, D, or E). Each entity is defined by a set of attributes: center $x$, center $y$, width, and height, forming a bounding box in the image. The logic is grounded in two multi-modal contexts: the text context ($T$), ``Potted Plant'', and the image context ($Img$), shown on the left. CALM \textit{truth evaluation} can calculate the truth value for each possible \assignment of the microwave: D receives a high score (0.8), C receives zero (0.0) because it is right of the sink, and E receives a low score (0.2) because it is left of the sink, but it is on the floor (less likely for a plant). A \textit{truth maximization} would choose D, the highest-scoring configuration. A \textit{truth-proportional sampling} might still occasionally select E (20\% of the time), allowing diversity in generative tasks, but it would never choose C.}
    \label{fig:method_example}
\end{figure}

CALM operates over entities that are composed of attributes. For example, in the kitchen scene shown in Figure~\ref{fig:method_example}, entities include the sink, dishwasher, and an entity ``plant'' for a new object; each is grounded by a bounding box defined by center $x$, center $y$, width, and height. 
We chose this representation for this application. However, other forms of grounding can be more suitable for other applications. 
Entities are related to each other through predicates, which yield analog truth values between 0 (completely false) and 1 (completely true). In the example, the logic relates the plant entity to the sink using two predicates: one semantic (\texttt{Category}) and one spatial (\texttt{Leftof}). These predicates are grounded with multi-modal context, allowing their truth to depend on real-world information. Contexts may include images, natural language text, or other sources that situate reasoning in the real world. In this case, an image of a kitchen (image context $I$) and the label "Potted Plant" (text context $T$). Predicates can form larger, more complex logical statements via connectives (e.g. $\land$, $\lor$, $\neg$) and quantifiers (e.g. $\forall_T$, $\exists_T$), allowing CALM to express rich, compositional logic. 
For example, in this figure, D is a probable grounding of the plant (hence receiving the highest truth value) compared to C and E. 
CALM allows for three types of inference on a given logic: truth evaluation, truth maximization, and truth-proportional sampling. Truth evaluation computes the truth value of a logic statement under a specific \assignment -- that is, a concrete set of values for all entities. In the example, evaluating the logic on candidate placement D yields a truth value of 0.8, while placements C and E yield 0.0 and 0.2, respectively. Truth maximization searches for the \assignment that makes the statement as true as possible, and would therefore select D. Truth-proportional sampling generates \assignments with likelihoods proportional to their truth values, occasionally selecting lower-scoring options like E for diversity in generation. Each predicate is backed by a predicate neural component, a trained model that guides toward the final truth value through a series of small refinements, using multi-modal context to describe which local choices are more aligned with the predicate. These components are trained by minimizing cross-entropy loss on real-data examples where the relationships expressed by the predicates are annotated or implicitly demonstrated.

\subsection{Defining CALM}
\input{tex/method_logic_def}

\subsection{Inference on CALM}
\input{tex/method_inference}

\subsection{Learning Multi-Modal Context for CALM}
\input{tex/method_learning}

%% file: tex/method_logic_def.tex
%We want to define a fuzzy logic that is grounded in multi-modal context. Images, text, etc can be used to semantically ground fuzzy predicates.
CALM extends classical logics by incorporating multi-modal data, such as images, text, and 3D models, into the evaluation of symbols and predicates. This provides semantic grounding from real-data contexts, and increases the expressiveness over purely symbolic representations.

\subsubsection{Entities and Contexts}

CALM represents knowledge using two forms: \textbf{entities} and \textbf{contexts}. An \textit{entity} is a symbolic object defined by a fixed set of attributes. Each attribute is a named component that holds a specific value or a domain of potential values. For example, in Figure~\ref{fig:method_example}, each entity corresponds to a physical object within an image, represented by a bounding box defined by four attributes: center~$x$, center~$y$, width, and height. An entity can either be a constant entity or a variable entity. 

A \textit{constant entity} has all attributes with fixed, known values. For instance, in the example, the sink and dishwasher are constant entities because their positions and dimensions are known and fixed in the given kitchen scene. Constant entities can be automatically grounded by linking their symbolic attributes to real-world perceptual data. For example, a YOLO object detector \cite{yolo} can identify objects already present in the scene and produce bounding boxes that specify their location and size within the domain of the image dimensions.

A \textit{variable entity}, on the other hand, has one or more attributes whose values are unknown and defined by non-singleton domains. The \text{plant} in the example is a variable entity: it does not yet have a fixed placement but instead has a domain consisting of multiple candidate placements (C, D, or E in Figure~\ref{fig:method_example}). 

\textit{Contexts} are external sources of perceptual or environmental information that ground CALM entities and predicates in the real world. Unlike classical logic, where meaning derives purely from abstract interpretation, CALM explicitly incorporates multi-modal contexts. Contexts can include images, natural language text, or other perceptual modalities, allowing logical evaluation to reflect real-world semantics and appearance. In Figure~\ref{fig:method_example}, CALM utilizes two contexts: an \textit{image context}, which provides visual data about the scene, and a \textit{text context}, containing the phrase ``Potted Plant’’.

\subsubsection{Predicates}
A \textit{predicate} in CALM is a logical function that takes as arguments one or more entities along with any number of contexts, returning a truth value in the interval $[0,1]$. Predicates define conditions or relationships among entities and contextual information. Predicates can be either bivalent or analog. A \textit{bivalent predicate} evaluates strictly to either 0 or 1, representing false or true based on fixed, hardcoded conditions defined by attributes of the entities. They behave as predicates in classical logic, without intermediate states. An \textit{analog predicate}, by contrast, returns an analog truth value between 0 (completely false) and 1 (completely true), allowing a nuanced evaluation based on multi-modal context. Some analog predicates are \textit{hybrid}, combining a hard component (that yields 0 or 1) multiplied by a soft component.

In the kitchen example from Figure~\ref{fig:method_example}, the predicates used are \texttt{Leftof(plant, sink; I)} and \texttt{Category(plant; ``Potted Plant'', I)}. Both are analog predicates, but \texttt{Leftof} is hybrid: it has a hard component that returns a truth value of 0 if the center $x$ of the plant entity is greater than or equal to the left edge ($x - \frac{w}{2}$) of the sink. This ensures that the plant is meaningfully to the left of the sink, not merely overlapping or adjacent. Otherwise, the soft component determines the analog truth value based on multi-modal context. In this example, a spot on the counter near the window will be more truthful for a potted plant than on the floor, while the opposite might be true for a chair. The \texttt{Category} predicate has no hard component and directly produces an analog truth value, indicating how well the entity matches the provided category description ("Potted Plant") given its location. 

Predicates operate over grounded entities. Before evaluating a predicate, all entities must have their attributes grounded: constant entities are given fixed values, and variable entities are associated with domains of potential values. Grounding happens at the entity level by mapping attributes to perceptual data, and at the predicate level by conditioning evaluation on multi-modal contexts such as images or text.

Each predicate type in CALM is defined by a name (notated $p$), an arity (the number of entities it relates), the specific attributes those entities must possess, and an \textit{affecting attribute set} $A_p$. The affecting attribute set specifies precisely which attributes influence the predicate's analog truth value. Returning to the example in Figure~\ref{fig:method_example}, \texttt{Leftof} takes two entities -- the plant and the sink -- and its truth depends only on their horizontal positioning: specifically, the center $x$ and width of both entities form its affecting attribute set, $A_{\text{leftof}} = \{x, width\}$. The \texttt{Category} predicate, which classifies a single entity, depends on all four bounding-box attributes: center~$x$, center~$y$, width, and height, since the perceived category may vary with size, position, and shape in the image.

Additionally, each predicate is associated with a predicate neural component $f_p$, a neural network which is used during inference to realize the grounding between multi-modal context and truth value; these components will be discussed in later sections.

In addition to \texttt{Leftof} and \texttt{Category}, we introduce several other analog predicates used throughout our experiments, including \texttt{Rightof}, \texttt{Above}, and \texttt{Below}. Each of these takes two entities with bounding-box attributes and an image context, and evaluates how strongly one entity is positioned relative to another in the corresponding spatial direction. While these predicates form the core set used in this work, CALM is compatible with a wide range of other predicate types that support diverse forms of multi-modal, context-sensitive reasoning.

%\subsubsection{Functions}

\subsubsection{Connectives}

CALM defines conjunction ($\land$), disjunction ($\lor$), and negation ($\neg$) as fundamental logical connectives. Each operates on predicates or statements, which return analog truth values in the interval $[0,1]$. CALM takes inspiration for these definitions from Zadeh fuzzy logic. Conjunction returns the minimum of its components' truth values ($A \land B$ is given by $\min(A, B)$). This captures the idea that a statement composed of multiple parts is only as true as its least true component. Disjunction returns the maximum of its components' truth values ($A \lor B$ is given by $\max(A, B)$). This allows the overall statement to hold if at least one of its parts is true to some degree. Negation returns $1 - A$, inverting the truth value of a single component. Full truth becomes falsehood, and vice versa, with intermediate values inverted accordingly.

\subsubsection{Quantifiers}

We define two quantifiers: universal and existential, parameterized by a threshold. The \textit{universal quantifier} requires that a predicate holds for all elements in a set, at least to a given analog truth threshold. Given a predicate $P(x)$ returning an analog truth value and a set $S$, the universal quantifier with threshold $T$ is defined as: $\forall_T x \in S, P(x) \equiv \min_{x \in S} P(x) \geq T.$  
This ensures that every element in $S$ satisfies $P(x)$ with at least analog truth value $T$. If any element in $S$ has a truth value below $T$, the statement evaluates to false (0); otherwise, it evaluates to true (1).

The \textit{existential quantifier} requires that at least one element in a set satisfies the predicate above a given threshold. It is defined as: $\exists_T x \in S, P(x) \equiv \max_{x \in S} P(x) \geq T.$  
This evaluates to true if there exists at least one element in $S$ for which $P(x)$ meets or exceeds $T$. Otherwise, it evaluates to false. This allows existential reasoning where only a sufficiently strong analog truth value is needed for satisfaction.

%% file: tex/method_inference.tex
%\subsection{Inference with CALM}

\begin{figure}[ptbh]
    \centering
    \includegraphics[width=0.85\linewidth]{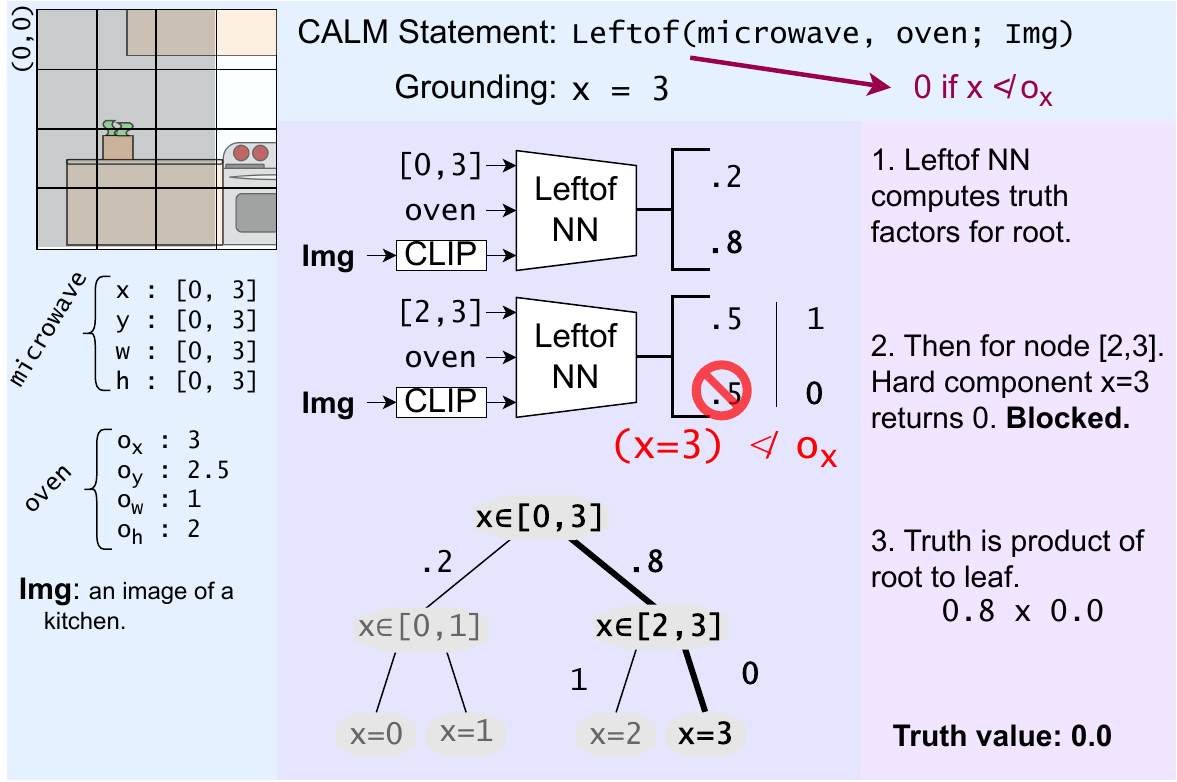}
    \caption{Truth evaluation for a single hybrid predicate, \texttt{Leftof(microwave, oven; Img)}. The microwave entity’s $x$ attribute is grounded to 3. At the root of its domain tree, the predicate’s neural component produces two truth factors for the left and right subdomains. Since the value 3 falls in the right child’s subdomain $[2,3]$, traversal continues to that node. There, the hard component of \texttt{Leftof} checks whether the microwave is strictly to the left of the oven -- specifically, whether its center $x$ is less than the oven’s left extent. Because both are 3, this condition fails, and the hard component returns 0. The predicate’s final analog truth value is thus 0, and no further traversal occurs.
}
    \label{fig:inftree1}
\end{figure}

\begin{figure}[ptbh]
    \centering
    \includegraphics[width=0.85\linewidth]{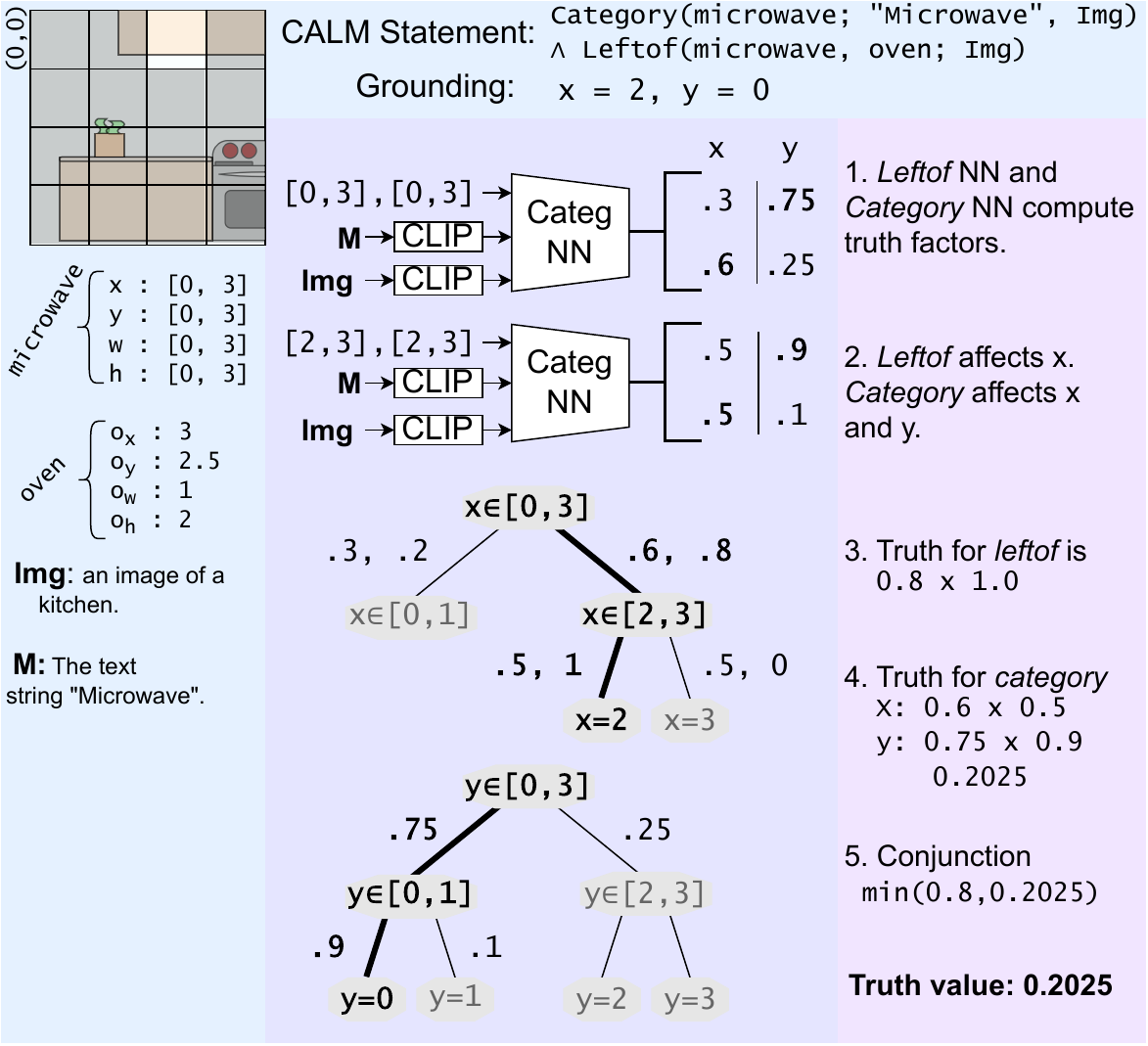}
    \caption{Extension of Figure~\ref{fig:inftree1}. This example evaluates a compound statement made of two analog predicates: \texttt{Category(microwave; "Microwave", Img)} and \texttt{Leftof(microwave, oven; Img)}. The microwave entity’s $x$ and $y$ attributes are grounded to 2 and 0, respectively. The \texttt{Leftof} predicate refines the microwave's $x$ (and $w$, but that is omitted here). It traverses the $x$ domain tree, producing a final truth of $0.8 \times 1.0 = 0.8$. The \texttt{Category} predicate is unary and takes two contexts -- an image and a text string. It refines all attributes, including $x$ and $y$ domain trees ($w$ and $h$ omitted). In the $x$ tree, the current subdomain is $[0,3]$, where the neural network outputs 60\% for $[2,3]$, then 50\% for $x=2$; the resulting truth factor is $0.6 \times 0.5$. In the $y$ tree, the same predicate outputs 75\% for $[0,1]$, then 90\% for $y=0$; this gives $0.75 \times 0.9$. The full truth value for \texttt{Category} is the product: $0.6 \times 0.5 \times 0.75 \times 0.9 = 0.2025$. The final truth of the statement is the minimum of its two predicate values: $\min(0.8,\ 0.2025) = 0.2025$.}
    \label{fig:inftree2}
\end{figure}

CALM includes three types of inference: truth evaluation, truth maximization, and truth-proportional sampling. Truth evaluation is the process of determining the analog truth value of a logical statement under a specific \assignment (complete specification of all unknown attributes for variable entities). This is useful for tasks involving scoring, verification, or consistency checking under fixed interpretations.

Truth maximization and truth-proportional sampling both involve generating \assignments for variable entities. Truth maximization searches for the \assignment $\alpha^*$ that yields the highest analog truth value under a given logic expression, making it suitable for tasks involving optimization, reconstruction, or controlled generation. Truth-proportional sampling instead draws \assignments with probability proportional to their analog truth values, enabling tasks that require generative modeling or probabilistic reasoning.

For example, in Figure~\ref{fig:method_example}, the microwave entity must be placed in the scene so that it satisfies the statement $\texttt{Category(microwave; T, I)} \land \texttt{Leftof(microwave, sink; I)}$. A truth evaluation computes how well each candidate placement (C, D, or E) satisfies the statement. A truth maximization selects the best-scoring candidate (D), while truth-proportional sampling will select E 20\% of the time and D 80\% of the time.

Each form of inference on CALM could be implemented in a variety of ways, with strengths and weaknesses depending on the setting and requirements. This work offers one. To perform inference efficiently, we use a data structure called a \textit{domain tree}. Each unknown-valued attribute of a variable entity is provided its own domain tree, which represents the attribute’s domain as a hierarchy of nested subdomains. The root node represents the full range of possible values, and each descendant partitions that space into progressively smaller subregions.

Inference can be implemented by traversing these trees through a process called \textit{refinement}. At each level, the inference algorithm selects one subdomain to continue exploring, gradually narrowing down the attribute's value. Refinement is driven by an interplay between the predicate neural network and the logic structure: the neural network provides multi-modal contextualization, while the logic enforces correctness and compositionality.

\subsubsection{Truth evaluation}
Given a CALM logic statement and a \assignment, Truth evaluation inference should return the analog truth value of all predicates.

\noindent\textbf{Domain trees.}  CALM represents the domain of each unknown-valued attribute of a variable entity using a hierarchical data structure called a \textit{domain tree}. Each attribute has its own $k$-ary domain tree, whose root node covers the attribute's entire domain. Every node in the tree partitions its domain into exactly $k$ child subdomains, each representing a distinct subset. For instance, with a binary tree ($k=2$) on a bounding box’s $y$ attribute, the left child might represent the low-valued top half of the image, and the right child the high-valued lower half. Leaf nodes correspond to single, fully specified values. Reaching a leaf node means assigning an exact value to the attribute.

\noindent\textbf{Truth factors.}  
This work calculates the analog truth value of predicates as a product of \textit{truth factors}. A truth factor is a number in $[0,1]$ that describes how truth is distributed across the $k$ subdomains at a given level of a domain tree. Truth factors are computed by the predicate neural component, which takes as input any multi-modal context and the current subdomain. It outputs $k$ truth factors; one for each child subdomain. It is analogous to a conditional probability, specifying how truth is distributed over subdomains given the current context. Truth values are computed as the product of truth factors along the relevant paths of the domain tree, just as joint probabilities are computed via the chain rule.

\noindent\textbf{Refinement.}  
Refinement is the process of narrowing down the domain of an unknown-valued attribute by selecting one subdomain at each level of its domain tree. This continues recursively until a leaf node is reached, determining a specific value for the attribute. In truth evaluation, the \assignment is already given, so the purpose of refinement is to identify the path through the domain tree that corresponds to the assigned value. At each step, the evaluator checks which of the $k$ children contains the value. For example, if the \assignment specifies that the width of an object is 12, refinement proceeds downward through the tree, selecting the subdomain that includes 12 at each level, until a singleton subdomain $\{12\}$ is reached.
Given a grounded logical statement and a complete \assignment of values to all variable entity attributes, truth evaluation proceeds as follows. First, for each predicate in the statement, the evaluator examines the relevant domain trees associated with the attributes involved. At each internal node along the path toward the assigned value, the evaluator checks whether the predicate’s hard component immediately returns 0 for the current subdomain. If so, the predicate evaluates to 0 and traversal halts. Otherwise, the evaluator invokes the predicate neural component, which consumes the multi-modal context and the current subdomain. The evaluator then selects the child node containing the assigned value and multiplies in the associated truth factor. After all refinement paths are traversed, the analog truth value of each predicate is given by the product of truth factors collected along the paths of its relevant attributes. The analog truth value of the full logical statement is then computed by applying logical connectives (e.g., conjunction as minimum, disjunction as maximum) to combine the individual predicate truths.

\noindent\textbf{Example.}  
This procedure is illustrated in Figure~\ref{fig:inftree1}, where the \texttt{Leftof} predicate is evaluated by refining the domain tree for the microwave entities horizontal position. During traversal, truth factors are generated at each node, but because the entity is not objectively left of the oven, the hard component enforces a truth value of 0. Figure~\ref{fig:inftree2} shows a complete statement evaluation involving both \texttt{Category} and \texttt{Leftof}. Each predicate is evaluated independently, and their analog truth values are combined using conjunction (minimum) to compute the final truth value for the statement.

\subsubsection{Truth maximization}

Truth maximization in CALM computes the \assignment $\alpha^*$, the grounding yielding the highest possible truth value for a logic statement. One way to implement this is through a brute-force truth evaluation of all possible \assignments -- this is not adequately efficient in most cases. Thus, this work presents a greedy-first refinement over domain trees followed by an exhaustive search. 
The idea is that greedy search usually returns a high-truth grounding, which can be used to prune unnecessary branches in the exhaustive  search. 
%associated with each unknown-valued attribute of the variable entities. Pruning is used to improve efficiency.

Initially, a greedy traversal explores each domain tree. At every node, the predicate neural components produce truth factors conditioned on multi-modal context, providing $k$ values corresponding to the $k$ child subdomains of the current node. Multiplying these new truth factors by the cumulative truth factors from previously traversed edges gives a value called the \textit{truth-so-far}, representing the overall truth accumulated along that partial refinement path. Each predicate independently calculates its own truth-so-far for each child edge relevant to its affecting attribute set. To select the next child node for traversal, the logic statement’s connectives amalgamate these predicate-specific truths-so-far into a single unified statement-level score. The child subdomain that maximizes this combined score is then chosen for greedy refinement.

If during traversal, the hard component of any predicate produces a truth factor of zero for a child subdomain, that path becomes \textit{blocked}. In such cases, if other child subdomains remain with non-zero truth factors, their truth factors are normalized to sum to one, allowing the traversal to continue with proportionality of remaining truth factors maintained. However, if all child subdomains of a node have truth factors of zero, the algorithm backtracks to the previous node, setting the truth factor leading into the blocked node to zero. This process follows the simple principle that a node's truth factor is zero if all of its children have zero truth.

Eventually, the greedy traversal reaches a leaf node, producing a first candidate grounding. However, the greedy path does not guarantee the absolute maximum truth value. Therefore, the truth value of this greedy leaf is saved as the current highest truth value, and the search continues as a backtracking depth-first search (DFS) with pruning. As DFS proceeds, any path with a truth-so-far less than the currently known highest truth value can be pruned immediately, as the product of truth factors monotonically decreases with deeper refinements. When the entire domain tree is either visited or pruned, the leaf with the highest truth value is returned as the grounding $\alpha^*$.

To illustrate, consider an attribute $x$ with domain $[1,4]$. The domain tree is binary and balanced, first splitting into subdomains $[1,2]$ and $[3,4]$, and then further into leaves $1$, $2$, $3$, and $4$. Suppose our statement is the conjunction $\texttt{rightof}(a,b) \wedge \texttt{category}(a,\text{``toaster''})$. At the root, predicates produce initial truth factors. Assume \texttt{rightof} produces truth factors of $0.9$ for subdomain $[1,2]$ and $0.1$ for $[3,4]$, and \texttt{category} produces $0.6$ for $[1,2]$ and $0.4$ for $[3,4]$. The conjunction (minimum) selects subdomain $[1,2]$ with a combined truth-so-far of $0.6$ (versus $0.1$ for $[3,4]$).

Next, subdomain $[1,2]$ splits into leaves $1$ and $2$. Suppose \texttt{rightof} now produces truth factors $0.8$ for leaf $1$ and $0.2$ for leaf $2$, while \texttt{category} gives $0.5$ for each leaf. Multiplying these by previous truth factors, \texttt{rightof} has a truths-so-far $0.72$ ($0.9 \times 0.8$) for leaf $1$ and $0.18$ ($0.9 \times 0.2$) for leaf $2$. Similarly, \texttt{category} yields a truths-so-far of $0.3$ ($0.6 \times 0.5$) for each leaf. The conjunction selects leaf $1$, as it has a higher combined truth-so-far ($0.3$) compared to leaf $2$ ($0.18$).

At leaf $1$, suppose the hard component of \texttt{rightof} evaluates to zero, blocking this leaf. Since this tree is binary, the truth factor for leaf $2$ of \texttt{rightof} is immediately normalized to $1.0$. Continuing with leaf $2$, we now have final truth factors from \texttt{rightof} as $1.0$ (after normalization) and from \texttt{category} as $0.4$. The truth-so-far for leaf $2$ thus becomes $0.18$ ($0.18 \times 1.0$) for \texttt{rightof} and remains $0.12$ ($0.3 \times 0.4$) for \texttt{category}. Taking the conjunction (minimum), leaf $2$ now has a truth value of $0.12$, recorded as the current best.

Backtracking, the algorithm revisits the previously unexplored subdomain $[3,4]$, which earlier had a combined truth-so-far of only $0.1$. As this truth-so-far is below the current best truth value of $0.12$, the subdomain $[3,4]$ is immediately pruned. Thus, leaf $2$ (attribute $x=2$) is ultimately returned as the grounding $\alpha^*$ for attribute $x$. Similar independent searches occur for other unknown attributes, yielding the final complete grounding.

\subsubsection{Truth-proportional sampling}

Truth-proportional sampling selects \assignments with a probability proportional to the analog truth value of a given logical statement. Similar to truth maximization, one straightforward method to achieve this is to evaluate every possible \assignment, calculate their truth values, and then normalize these values to form a sampling distribution. While this brute-force approach is conceptually simple, it quickly becomes computationally infeasible as the number of \assignments grows. This inefficiency arises fundamentally because analog logic statements composed of connectives such as conjunction (min) and disjunction (max) inherently disrupt any straightforward probabilistic structure. As a result, efficiently sampling proportionally from a single predicate is significantly easier than sampling from a more complex, compound statement. This work offers an efficient exact sampling algorithm for the former (inspired by SampleSearch \cite{samplesearch}), and an efficient approximate algorithm for the latter.

\noindent\textbf{Sampling a single predicate.} 
Truth-proportional sampling from a single predicate can be efficiently accomplished using ancestral sampling over its associated domain tree. At each node within the domain tree, sampling is guided by the truth factors provided by the predicate's neural component. Recall that these truth factors are produced by neural evaluations conditioned on multi-modal contexts, and they are analogous to conditional probabilities: each truth factor represents the proportion of truth distributed to each of the node's child subdomains given that the parent subdomain has already been selected.

This sampling process begins at the top of the domain tree. At the root node, the predicate neural component examines the entire possible range of attribute values and computes a truth factor for each child subdomain based on context information. A reminder, these truth factors are numbers between 0 and 1, with all children summing to 1, showing how strongly each subdomain satisfies the predicate. Therefore, one child subdomain can be randomly selected, with higher truth factors corresponding to higher chances of selection. After selecting a child, the procedure repeats at the next level down, again computing new truth factors for that child's subdomains and randomly selecting one according to these truth factors. This continues downward until reaching a leaf node, which specifies an exact attribute value.

As with truth maximization, blocking may occur if the hard component of the predicate evaluates a particular child subdomain to zero truth. In such a case, the blocked subdomain is immediately discarded. Truth factors of the remaining child subdomains are renormalized to ensure they sum to one, thus maintaining correct proportionality for all non-blocked branches. This blocking and renormalization process follows the principle established by SampleSearch~\cite{samplesearch}.

As an example, consider a single attribute domain tree with two levels, initially splitting into subdomains $[1,2]$ and $[3,4]$, which further split into individual leaves. Suppose at the root, truth factors computed by the predicate neural component are $0.9$ for the left child $[1,2]$ and $0.1$ for the right child $[3,4]$. Sampling selects subdomain $[1,2]$ with $90\%$ probability and subdomain $[3,4]$ with $10\%$ probability. Assuming subdomain $[1,2]$ is chosen, the process repeats for its child nodes, say leaf $1$ and leaf $2$. If the neural component assigns truth factors of $0.75$ and $0.25$ respectively, leaf $1$ is chosen with $75\%$ probability and leaf $2$ with $25\%$ probability, given that $[1,2]$ was selected. If no blocking occurs, this entire procedure requires only a single traversal from root to leaf, making it highly efficient. However, this efficient method cannot straightforwardly extend through logical connectives.

\noindent\textbf{Sampling a full statement.}
Sampling proportionally to the analog truth value of compound statements built from connectives remains a challenging area. The difficulty arises because connectives disrupt the tree-based probabilistic structure that makes predicate-level sampling straightforward. Unlike a single predicate, compound statements combine multiple predicates through non-linear operations such as conjunction (min) or disjunction (max), resulting in complex and non-factorizable sampling distributions.

As an example of this problem, consider a binary $k=2$ domain tree over a single attribute $x$. The statement being evaluated is two predicates $P_1$ and $P_2$ connected by a conjunction ($\land$, implemented with min). $P_1$ and $P_2$ are over the same entities, and both contain $x$ in their affecting attribute set, so both ``have a say'' in the refinement of $x$ to get a truth value. $P_1$ has 0.75 and 0.25 as truth factors at the root node, while $P_2$ has 0.9 and 0.1. on the left child, they have 0.2, 0.8 and 0.6, 0.4 respectively, all leading to leaves. on the right,  they have 0.7, 0.3 and 0.5, 0.5 respectively, also leading to leaves. Either $P_1$ or $P_2$ can be sampled efficiently using the method described in the previous section. For $P_1$, select left with 75\% chance and right with 25\%, then continue in the next node. But how can these two \textit{channels} of truth factors be combined onto one domain tree? One possibility is to take the truth factors of both at each node, and combine them through the connective's logic. The root node would then have factors $min(0.75, 0.9)$ and $min(0.25, 0.1)$. But this does not produce a valid distribution (0.75 and 0.1 do not sum to one), and is even if they were renormalized, the product of these would not result in a distribution proportional to truth -- the leftmost leaf would have a truth value of roughly $0.2941$, versus its true value of $min(0.75 \times 0.2, 0.9 \times 0.6) = 0.15$.

A way of getting around this is with a brute force approach. For each predicate, the truth value of all possible groundings can be evaluated by calculating every root-to-leaf product for its own truth factors. Then, once complete truth values exist for every grounding of the predicate, they can be calculated for the entire statement by applying the connective logic. The truth values can then all be normalized to create a distribution proportional to truth, and sampled. Once the brute force algorithm has been run on a statement, it can be sampled in constant time, but this does little to detract from the infeasibility of calculating truth for every possible grounding.

To compensate for the inefficiency of the brute force approach, this work introduces an approximate sampling algorithm based on the previous predicate proposal sampling followed by truth-proportional resampling. In this approach, each predicate independently generates candidate \assignments using its own efficient truth-proportional sampling method described above. After collecting these candidate \assignments, each candidate is evaluated against the entire compound statement. Specifically, this means computing the truth values of all predicates involved at each candidate \assignment and then combining these values according to the connectives. This evaluation results in an analog truth value for the full logical statement at each candidate \assignment. Next, these statement-level truth values are normalized to form a proper sampling distribution across all candidate \assignments. Finally, one candidate \assignment is selected from this distribution, ensuring approximate proportionality to the statement's truth.

As a concrete illustration, consider a compound statement composed of two predicates combined by conjunction: $\texttt{Leftof}(a,b)$ and $\texttt{Category}(a, \text{``Microwave''})$. First, we independently sample multiple candidate \assignments from each predicate. Suppose one candidate \assignment sampled from the predicate $\texttt{Leftof}$ yields a high truth value of $0.6$ for that predicate. Another candidate \assignment sampled independently of the predicate $\texttt{Category}$ has a high truth value of $0.8$ for the $\texttt{Category}$ predicate. Note, these two candidates may differ substantially, since each predicate's sampling distribution depends only on itself.

We then evaluate each candidate \assignment using the entire compound statement. For the first candidate \assignment, we evaluate both $\texttt{Leftof}$ and $\texttt{Category}$ predicates. Suppose evaluation yields predicate truth values of $0.6$ (for $\texttt{Leftof}$) and $0.8$ (for $\texttt{Category}$). Since the statement uses conjunction, the overall truth value is the minimum: $0.6$. For the second candidate \assignment, assume evaluation gives both predicates a truth value of $0.8$, so the full statement's truth is $0.8$. These resulting statement-level truth values ($0.6$ and $0.8$) are normalized to form sampling probabilities: $0.6 / (0.6 + 0.8) = 0.43$ and $0.8 / (0.6 + 0.8) = 0.57$. Finally, one of these candidate \assignments\ is selected according to these probabilities.

While practical, this predicate proposal sampling and resampling method only approximates truth-proportional sampling for compound statements. Efficient exact truth-proportional sampling from arbitrarily complex CALM statements remains an open limitation of this work.

%% file: tex/method_learning.tex
%\xyx{Learning for CALM learns the truth values of predicates in a given context.} 
%What is the input... What is the output...
%For example, ..., 

%\xyx{Leveraging the notion of domain trees, the learning task corresponds to learning the what?? values along each branch...}

%\xyx{On a high level, how is this learning achieved?} Translate into a sequential prediction task? 

%\bigskip

%\xyx{======= Original writeup ========}
Each predicate type $p$ in CALM is associated with a predicate neural component $f_p$, which must be trained to produce meaningful truth factors. The goal of learning is to train $f_p$ so that it assigns high truth factors to subdomains that lead toward correct \assignments. Training is fully supervised.

A training example consists of a grounded predicate instance, a multi-modal context, and a ground-truth \assignment $\alpha_{\text{gt}}$. The \assignment $\alpha_{\text{gt}}$ maps each unknown-valued attribute of a variable entity to a specific value. For every predicate instance, the attributes involved are those in its affecting attribute set $A_p$.

The input to $f_p$ includes the attributes of all entities that are argument to the predicate and the contextual feature vectors. The attributes are represented as 2 inputs: the min and the max of the current subdomain. If an attribute is known (as in a constant entity), its min and max are equal. The neural network concatenates all attribute representations and context vectors to form its input.

For each attribute in $A_p$, the network outputs $k$ truth factors -- one for each subdomain at the current node in the domain tree. These outputs are produced by softmax-activated heads, one per attribute.

The loss is computed by comparing the predicted truth factors against the correct decision, which is extracted from the path that the ground-truth \assignment takes through the domain tree. For example, if we are given an annotated image where a microwave is to the left of an oven, we treat the microwave’s position as ground truth and decompose it into a sequence of domain tree decisions. Each decision is a classification among $k$ subdomains. The predicted truth factors define a probability distribution over those options. We apply a \textbf{cross-entropy loss} between the predicted distribution and the correct subdomain choice. This loss is used to update the parameters of $f_p$ through gradient descent.

%\subsubsection{Multi-modal Feature extraction} 
A key consideration in training CALM predicates is embedding multi-modal contextual data into vectors usable by each neural component $f_p$. In other words, how do we get the contextual feature vectors that facilitate grounding in real-world multi-modal data? This can be achieved through various methods, but in this work we utilize a pretrained CLIP model\cite{radford2021learning,ilharco_gabriel_2021_5143773}, which encodes image and text contexts into a shared latent embedding space. Specifically, context inputs like a background image and a descriptive text string (e.g. defining the category of a bounding-box object) are run through CLIP’s encoders, producing a 512-dimensional latent vector. This vector can then serve as input to any downstream predicate neural networks. This procedure is visualized in Figures~\ref{fig:inftree1} and~\ref{fig:inftree2}, where CLIP embeddings of the background image and text description form the basis for predicate evaluations.

%% file: tex/experiments.tex
\begin{figure}[tb]
    \centering
    \includegraphics[width=0.8\linewidth]{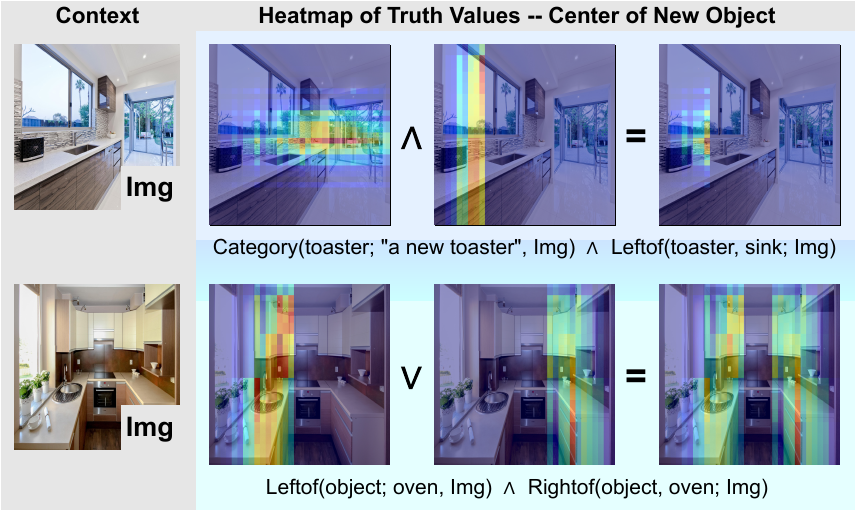}
    \caption{CALM maintains the composability of classical logic. Above, heatmaps are generated using CALM predicates. Hotter pixels have higher truth values for being the center position of a new object (\textit{toaster} and \textit{object}). By combining predicates with conjunctions (top) or disjunctions (bottom), more complex ``truthscapes'' can be formed from simple units. }
    \label{fig:heatmap_plate_1}
\end{figure}

Formal logic systems offer structure, precision, and interpretability. However, they lack semantic grounding in perceptual context. Neural networks, by contrast, ground concepts in real-world multi-modal data but sacrifice formal structure and logical compositionality. CALM bridges this gap by embedding multi-modal neural modules directly within an analog logic framework, enabling structured reasoning over grounded, perceptual inputs. To evaluate whether CALM effectively unifies symbolic reasoning with perceptual grounding, this work offers a set of experiments focused on spatial reasoning and object placement under ambiguity. Each experiment tests CALM’s ability to interpret and apply logic in context. 

The first experiment presents occluded visual scenes (e.g. a kitchen scene) alongside spatial logic (e.g. “the chair is right of the sink”) and asks models to infer where missing objects belong. By using learned clues from within the scene combined with hard logical inference, CALM consistently outperforms both a vision-language model baseline and a formal logic baseline, especially under partial information.

The second experiment tested whether CALM’s predicate truth scores behaved meaningfully across the full domain of possible object placements. Logic statements were converted into heatmaps, where each pixel represented the truth value of placing the object there. CALM produced structured, interpretable heatmaps that aligned with spatial intent, unlike the neural baseline. A human study confirmed CALM’s outputs were significantly more aligned with the logic ($p < 0.0001$).

The final experiment demonstrated that CALM’s truth-proportional sampling could guide object insertion in generative models. By sampling placements proportional to truth and passing them to Stable Diffusion, CALM enabled realistic, diverse, and logic-consistent image inpainting.

\begin{figure}
    \centering
    \includegraphics[width=\linewidth]{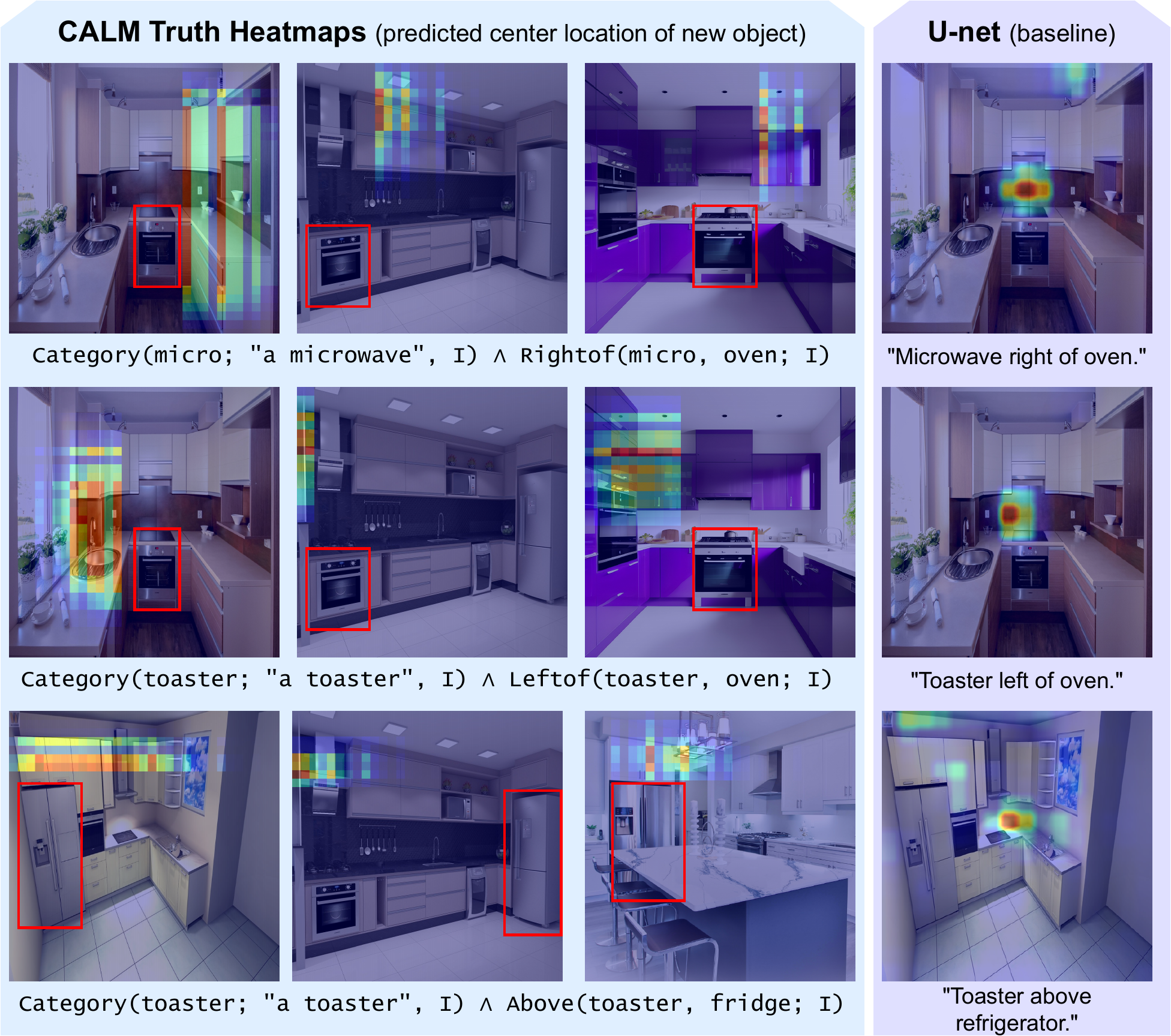}
    \caption{A comparison of heatmaps generated by CALM and a U-net baseline. While both are capable of dynamic, accurate predictions as to where a new object should be placed in the scene, CALM displays a much clearer alignment with the spatial relationships. Both methods are well-grounded in multi-modal data, but CALM retains the ability to also apply abstract reasoning, ensuring its prediction aligns with its logic. The U-net is softly conditioned by its input text, but lacks explicit logical structure, leading to unaligned predictions in each of the images given. When presented with similar images in a human study, participants showed significant preference for CALM's heatmaps (see Section~\ref{sec:human_study}).}
    \label{fig:heatmap_compare}
\end{figure}

\begin{figure}
    \centering
    \includegraphics[width=\linewidth]{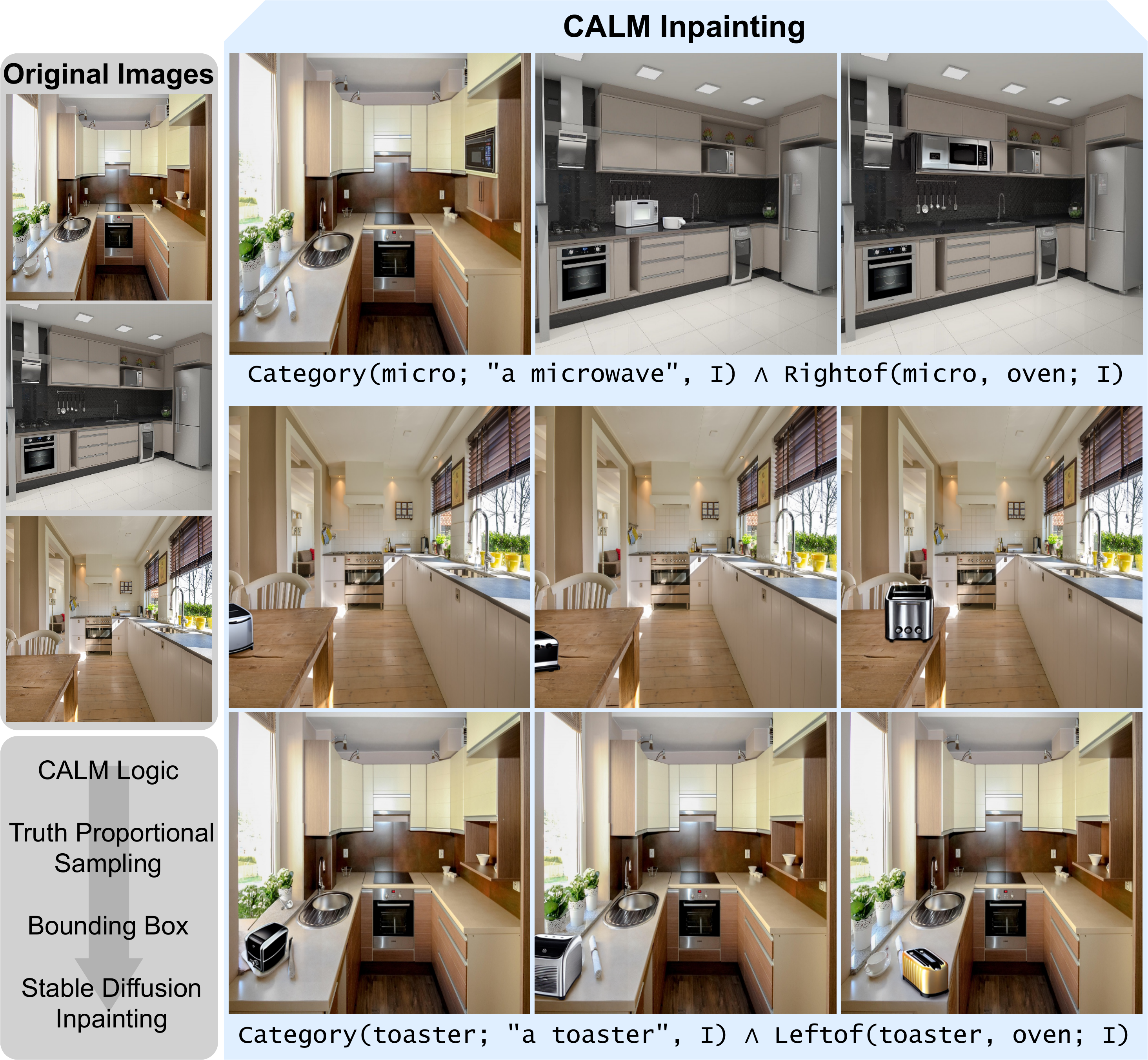}
    \caption{Examples from the CALM inpainting demonstration. CALM is capable of sampling bounding boxes proportional to the truth value of the given grounded logic. These boxes can then be given to an off-the-shelf diffusion model like Stable Diffusion \cite{rombach2021highresolution} to actually add these objects to the image. This approach is inspired by the SPRING system \cite{spring}.}
    \label{fig:inpaint_plate}
\end{figure}

The experiments provide the following key takeaways:
\begin{itemize}
\item CALM achieves 92.2\% accuracy in spatial fill-in-the-blank tasks with partial logic, outperforming FOL (86.2\%) and LLM baselines (59.0\%) (see Figure~\ref{fig:fitb_results}).
\item Heatmaps generated from CALM logic match human expectations significantly better than a neural baseline (see Figure~\ref{fig:heatmap_compare}).
\item Truth-proportional sampling enables good-looking, constraint-respecting image edits when paired with generative models (see Figure~\ref{fig:inpaint_plate}).
\end{itemize}

\subsection{Testing Truth Evaluation with Fill-in-the-blank Scenes}
\label{sec:fitb_section}

\subsubsection {Setup}
The task consisted of assigning objects to occluded regions in an image using a provided logic description. Each input included an image with blanked-out object regions, a list of candidate object descriptions (e.g., ``cup’’ or ``bowl’’), and a textual statement expressing spatial relationships among them (e.g., ``the cup is left of the bowl’’). The model was required to produce a mapping from object descriptions to blank locations such that the logical constraints were best satisfied in context. The locations should also satisfy common sense. 

To vary the amount of information available, each trial specified a different percentage of the total possible spatial relationships: 0\%, 50\%, or 100\%. At 0\%, the model received no spatial logic, just object categories. At 50\%, each possible relationship (e.g., left/right, above/below) had a 50\% chance of being included. At 100\%, all pairwise spatial relationships were present in the text.

The primary evaluation metric was \textit{object accuracy} -- the percentage of objects assigned to the correct locations, as defined by the original annotations. In addition, \textit{scene accuracy} was measured as the percentage of scenes in which all object placements were correct. For example, if a scene contained four objects and the model placed one incorrectly, the object accuracy for that scene would be 75\%, but the scene accuracy would be 0\%, since some objects were placed incorrectly (in this case, 1).

\noindent\textbf{Dataset.} For training and testing, images were drawn from the COCO 2017 dataset \cite{COCO}. Scenes were filtered to only include those depicting indoor objects: chairs, couches, potted plants, beds, mirrors, dining tables, desks, microwaves, ovens, toasters, sinks, refrigerators, clocks, and blenders. Scenes were resized and padded to a uniform resolution of 128 by 128 pixels. For each selected image, two or more objects were removed, and their bounding boxes were used as candidate locations. These regions were inpainted using the Telea inpainting \cite{telea2004image}, creating visually plausible occlusions without revealing object identity.

\noindent\textbf{CALM Agent.} For each scene, CALM received an inpainted image, a list of object labels (e.g., ``microwave,'' ``toaster''), and a text-based spatial logic description (e.g., ``the microwave is left of the oven''). Each blank region was represented as a structured variable with four components: x-center, y-center, width, and height. CALM constructed a logic expression consisting of category predicates for each object and spatial predicates for any specified pairwise relations. The system evaluated all injective mappings from objects to blank regions and selected the one that yielded the highest overall truth value for the logic expression.

Each predicate used one or more neural components to guide truth estimation during variable refinement. The spatial predicates (``left of,'' ``right of,'' ``above,'' ``below'') refined related attributes -- either x-center and width or y-center and height -- and produced four truth factors corresponding to binary domain trees ($k=2$). The category predicate refined x-center, y-center, width, and height, producing eight truth factors in total for four trees.

Each spatial predicate network received a concatenation of position coordinates and a ResNet18 \cite{he2016deep} image embedding. This was processed by a three-layer fully connected network ($\text{input} \rightarrow 128 \rightarrow 64 \rightarrow 4$) with batch normalization, dropout (0.5), and leaky ReLU activations. The category predicate network used ResNet18 to encode the image and a projection of the CLIP text embedding (pretrained only, with ViT-B-32 parameters \cite{dosovitskiy2020image}). These were combined with normalized domain data (the domain at the current node) and passed through a two-layer feedforward network ($\text{input} \rightarrow 64 \rightarrow 8$) to predict spatial refinements.

\noindent\textbf{CALM Training.} All predicate networks were trained using supervised learning on the COCO training data. For spatial predicates, positive pairs of objects were selected based on ground-truth relationships, and the correct direction of refinement was determined from their relative positions. For the category predicate, individual labeled objects were used, with ground truth targets derived from their annotated bounding boxes. Networks were trained to predict these decisions using cross-entropy loss, pushing them to learn the distribution of refinements. Models were optimized with Adam \cite{adam} and trained for 200 epochs (chosen from visual inspection of COCO validation loss curve) with a batch size of 128 (32 and 64 were also tried). A learning rate of $2 \times 10^{-4}$ was used. A range from $1 \times 10^{-3}$ to $1 \times 10^{-5}$ was tested, with $1 \times 10^{-4}$ showing comparatively good results, albeit with slowed convergence. A modest GPU (NVIDIA GeForce RTX 2070 Super) was utilized to speed up training. 

\noindent\textbf{LLM Baseline.} This baseline used the GPT-4o \cite{openai2024gpt4omini} vision-language model to assign objects to blank regions, given an image and a spatial logic description. The image was first inpainted using the Telea method (as with CALM's data), and each blank region was outlined in a distinct color. These colors, along with the center coordinates and size of each blank, were included in the prompt. Each object was listed with an index and name (e.g. ``0: microwave''), and the logic statement was appended exactly as used in the CALM agent. The agent then used Chain-of-Thought \cite{cot} reasoning to come to a solution over several steps. The model was queried via API with no fine-tuning or additional examples, operating in a fully zero-shot setting.

\noindent\textbf{FOL + Uniform Sampling.} This baseline used the same logical structure as CALM but operated with a classical first-order logic. All predicates were treated as bivalent: an object-to-blank assignment either satisfied the logic or it did not, with no grounding in the context of the image or object descriptions. The system sampled uniformly from the set of satisfying assignments. This baseline isolated the contribution of symbolic reasoning alone, without any perceptual grounding or analog truth values.

\subsubsection{Results}

\begin{figure}
    \centering
    \includegraphics[width=\linewidth]{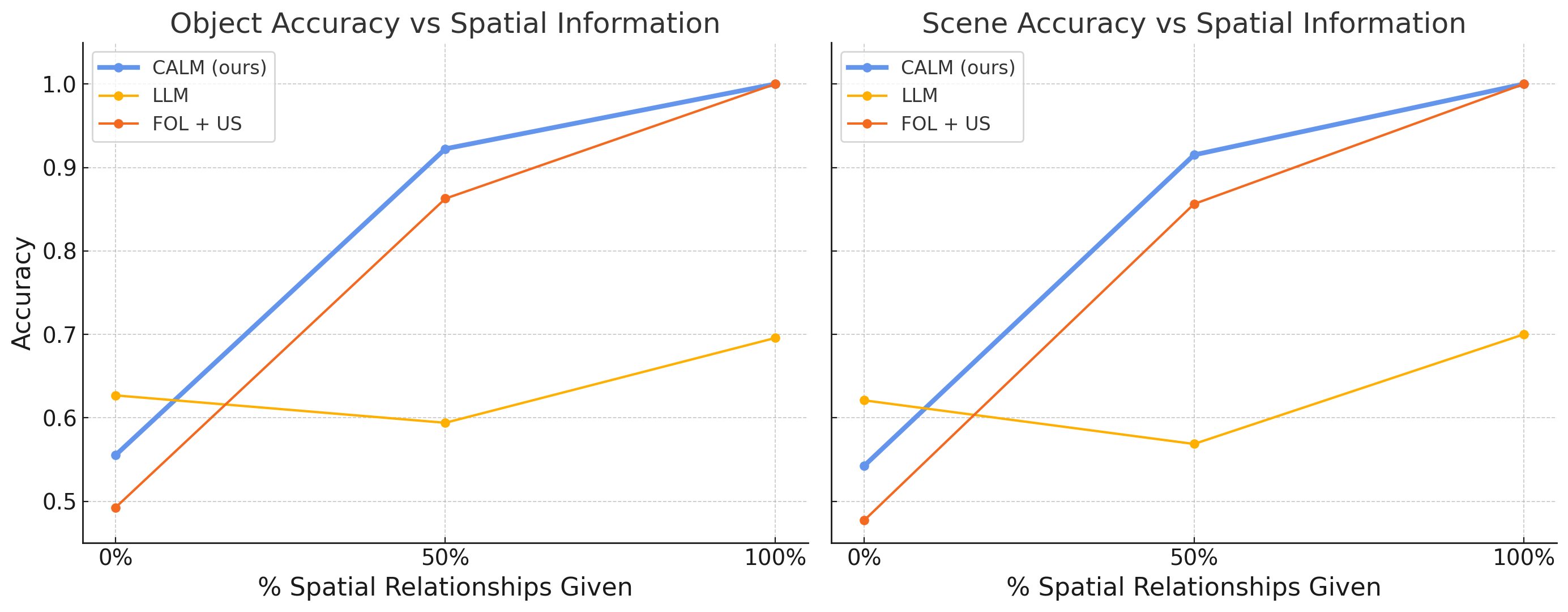}
    \caption{Results from the fill-in-the-blank experiment. Object accuracy and scene accuracy metrics are shown as a function of spatial logic completeness. CALM improves smoothly with more logical constraints, while FOL + Uniform Sampling relies heavily on sufficient logic to succeed. The LLM baseline performs inconsistently, showing weak alignment with the given spatial structure.}
    \label{fig:fitb_results}
\end{figure}

Figure~\ref{fig:fitb_results} shows accuracy for each method across varying amounts of spatial information. CALM achieves 55.5\% accuracy with no spatial logic, rising sharply to 92.2\% at 50\% and reaching 100\% when fully specified. The FOL + Uniform Sampling baseline follows a similar yet lower-accuracy trend, from 49.3\% to 86.3\% to 100\%. The LLM baseline performs better than both methods at 0\%, with 62.7\% accuracy, but degrades to 59.4\% at 50\% and only modestly improves to 67.8\% at 100\%.

These results support the hypothesis that CALM produces more semantically aligned truth evaluations than symbolic or neural methods alone. The strong performance of the FOL baseline at 50\% reflects the relatively low complexity of typical COCO scenes -- often just two or three objects -- where even partial logic can tightly constrain the solution space. However, without perceptual grounding, this approach fails catastrophically in more

\subsection{Human Study with Heatmaps}
\label{sec:human_study}

\subsubsection {Setup}
By calculating the truth value for every grounding of a bounding box in an image, a heatmap can be created where warmer colors represent a higher truth for the grounding (see Figure~\ref{fig:heatmap_compare}). This gives a complete view of the spatial understanding of a scene. This work evaluates CALM's spatial reasoning and contextual understanding by comparing these heatmaps with a U-Net-style encoder-decoder baseline in a human study, where participants decide which heatmap of a scene they find better aligned with the logic and the background image.

\noindent\textbf{Dataset.} This study used the same training data as Section~\ref{sec:fitb_section}. Background images for the test samples showed to the participants were sourced online (this dataset will be released with this work), forming a small dataset of 10 unique backgrounds. Fitting logic specifications were then assigned to each. This manual approach for the test data was necessary to create believable logic-background pairs.

\noindent\textbf{CALM Heatmap Generator.} CALM produced heatmaps by evaluating the truth value of every pixel as a center point to a bounding box. All variables were given a domain between 0 and 31 to accommodate a 32x32 heatmap grid, which could then be overlaid on an image via bilinear upscaling.

\noindent\textbf{CALM Training.} This study used the same training procedure as Section~\ref{sec:fitb_section}.

\noindent\textbf{U-net Baseline.} As a baseline to represent non-logic-driven neural approaches at heatmap generation, this work utilizes a text-conditioned U-Net-style encoder-decoder. The image backbone consists of repeated blocks of two 3x3 convolutions, each followed by batch normalization and ReLU activation, with three successive 2x2 max-pooling operations for downsampling. A bottleneck block applies the same double-convolution structure at the lowest resolution. The decoder reduces channel dimensionality using a 1×1 convolution and upsamples the spatial resolution to a fixed 32x32 using adaptive average pooling. The text embedding is processed by CLIP, then projected to match the image feature channel dimension through a linear layer, broadcast across spatial dimensions, and combined with the image features using a channel-wise dot product.

\noindent\textbf{U-net Training.} The U-net-style model was trained for 10 epochs (chosen from visual inspection of COCO validation loss curve) using a batch size of 8. Ground-truth heatmaps were generated from binary object masks and downsampled to 32x32 to align with the model’s output resolution. The loss function was a distributional cross-entropy, treating each heatmap as a spatial probability distribution by normalizing the ground-truth mask and comparing it to the model’s softmax-normalized output. The CLIP model was not optimized, using the same pretrained parameters as CALM \cite{dosovitskiy2020image}, but the projection layer and the rest of the encoder-decoder were. Optimization was performed using Adam with a fixed learning rate of $1 \times 10^{-4}$ --  $1 \times 10^{-3}$ and  $1 \times 10^{-5}$ were also tested during tuning

\noindent\textbf{Study Structure.} Each participant was asked to rate 10 pairs of images. Each pair used the same background scene and same logic (which is displayed as natural language text below both images). The participant was asked to select on a scale 1 through 5 -- ``Choosing lower numbers favors the left side, while choosing higher numbers favors the right. Choosing the middle means both images are equally good/bad. Or no clear preference''. Which algorithm was assigned to which side was randomized for each pair. 

Two metrics were collected through two of these rating questions. \textit{Text alignment preference} was collected with the question ``Which image is more accurate to the text? Which one fits the sentence better? If you put an object in one of the warm color areas, would it fit the sentence?'' It represents how well a truth mapping aligns with the logic that produced it, relative to the simple text conditioning of the neural network. \textit{Background alignment preference} was collected with the question ``Which image is more realistic for the background image? Which positions look more natural in the scene? Like an object of this type would be installed there. If you put an object in one of the warm color areas, would the kitchen look realistic?'' It represents how well a truth mapping aligns with the perceptual context the logic is grounded in (in this case, the background image).

Participants were given a short tutorial before beginning the survey, giving them a clear example with handcrafted heatmaps where one was clearly more correct in both metrics than the other. Two gold standard questions were employed to track rushed or unconsidered answers.

\noindent\textbf{Demographics.} 24 participants were surveyed. One admitted to not understanding the survey and was phased out, leaving 23 active participants. These participants ranged in age from 18 to 60 years old, and had a mix of education levels from high school diploma up to graduate degree. 18 reported completing the survey from a phone screen, 5 from a laptop, and 1 from a desktop. 8 participants reported normal vision proficiency, while 16 reported corrected-to-normal vision (glasses or contacts). None reported color vision deficiency or other visual impairment.

\subsubsection {Results}
Some examples of our CALM heatmaps can be seen in Figure~\ref{fig:heatmap_compare}. The human study showed that CALM significantly outperformed the baseline in aligning heatmaps with the provided textual logic. Participants rated CALM-generated heatmaps at an average of $3.76 \pm 1.38$ (on a scale where above 3 indicates a preference for CALM), with high statistical significance ($t = 8.34$, $p < 0.0001$). This clearly demonstrates that CALM's approach produces spatial reasoning heatmaps that participants find markedly more consistent with explicit textual descriptions.

In contrast, when evaluating alignment with the background scenes, CALM performed similarly to the U-net baseline. CALM heatmaps had an average rating of $3.10 \pm 1.45$, indicating only a slight preference, but without achieving statistical significance ($t = 0.998$, $p = 0.16$). This implies that CALM and the baseline are roughly equivalent in how naturally their object placements fit within the background scene, as judged by human perception of spatial affordance and environmental plausibility.

\textbf{This suggests that CALM is highly effective at logically grounding spatial reasoning}, and roughly equivalent to standard neural models in perceptual contextualization. It offers strong interpretability and logical consistency without sacrificing human-aligned placement in scenes. CALM’s ability to separate semantic reasoning from raw visual modeling makes it well-suited to serve as a reasoning backbone that integrates cleanly with generative models or downstream decision-making.

\subsection{Demonstrating Truth Proportional Sampling Scene Inpainting}

Like many logical systems, CALM was well suited for high-level reasoning over downstream tasks. We demonstrated this by integrating the CALM spatial reasoning measured in the previous experiments into image generation models (see Figure~\ref{fig:inpaint_plate}). This approach took inspiration from recent work in controllable image generation with diffusion models \cite{spring,zhang2023adding}.
The goal of object inpainting was to synthetically generate realistic instances of an object in a designated region of an image. With CALM, this goal expanded to adding images according to an analog logic proposed by a human designer.

\subsubsection{Setup}
A human designer defined a logical expression describing a desired spatial relationship (e.g., placing "a microwave right of the oven"). CALM then performed truth proportional sampling to generate realistic bounding boxes satisfying this logic. By sampling proportionally to truth, the placements produced were diverse yet guaranteeably constrained by the specified logic and aligned realistically with the scene context. These bounding boxes were automatically passed to the diffusion model, specifying the precise region within the image for object placement.

\noindent\textbf{Integration with Diffusion Model.} Once bounding boxes were sampled by CALM, they were passed to the diffusion model -- in this case Stable Diffusion \cite{rombach2021highresolution}. Models like these were naturally capable of realistic inpainting -- that is, writing over a portion of the image defined by a mask, and doing so in a way that appeared natural after the image was edited.

The model was prompted with the category string (e.g. "a microwave" or "a toaster"), and was applied over square crops centered on the target bounding box, resized to 512x512. This crop was done because Stable Diffusion had trouble generating objects within relatively small masks. Each crop was padded with an additional margin (32 pixels) to ensure that the diffusion model had enough surrounding context to fit the new object in naturally. The bounding box defined the inpainting mask used, enforcing the newly generated object to exist only within the bounding box.

Inpainting was run with 100 inference steps and a guidance scale of 7.5. After inpainting, the generated region was rescaled and pasted back into the original full image frame.

\subsubsection {Results}
Figure~\ref{fig:inpaint_plate} demonstrats CALM’s ability to generate diverse yet logically consistent object placements for two distinct logic statements across nine images. Notably, the sampled locations and visual appearances varied significantly due to the sampling, yet they always remained highly realistic and faithfully adhered to the spatial constraints specified by CALM’s analog logic.

This result would be improbable for purely neural approaches, which frequently violate logical constraints and lack guarantees for consistency. It would likewise be impossible for purely bivalent classical logic methods, which cannot express nuanced spatial preferences or leverage detailed perceptual context. CALM uniquely bridged this gap, delivering both the expressiveness needed for contextually-grounded interior design preferences and the strict logical structure necessary to ensure consistency.

%% file: tex/conclusion.tex
This work introduced Contextual Analog Logic with Multimodality (CALM), a logic that operates over analog truth values and that embeds multi-modal contextual information directly into predicate evaluation through neural networks. 
CALM allows us to reason about intricate 
human preferences in complex environments. 
We began with a key problem: classical logic is powerful but limited -- it can only handle binary truth values and lacks the grounding needed to interpret symbols in real-world settings. 
Though previous fuzzy logic improved expressiveness, and neural networks are capable of extracting multi-modal information, neither provides a logic system capable of both. 
Existing approaches fail to produce logic that is expressive enough, while retaining the compositional structure needed for complex reasoning. CALM addresses this by creating a truth-graded analog logic with neural grounding, allowing truth values to reflect contextual perception while supporting structured, symbolic inference.

CALM supports three types of inference: truth evaluation, truth maximization, and truth-proportional sampling. These enable it to not only verify whether a statement holds in a given context, but also to generate optimal or plausible assignments of variables in tasks like spatial placement or image completion. This makes CALM suitable for both analysis and generation in perception-grounded reasoning tasks.

%Despite its strengths, CALM has limitations. Truth-proportional sampling over large, connected logical expressions can be computationally expensive, forcing a tradeoff between exhaustive search and approximate sampling. Furthermore, CALM still requires hand-designed predicate structure and cannot yet learn logical topology end-to-end. These constraints limit its scalability in domains where logic is less explicit or more entangled.

The experiments in this work show CALM’s effectiveness in spatial reasoning tasks. In a fill-in-the-blank scene completion task, CALM achieved 92.2\% object accuracy with only partial information, outperforming classical logic by 6 percentage points and a vision-language model by 33. In a human evaluation of heatmaps, CALM outperformed a neural baseline by better aligning with logical instructions with high significance ($p < 0.0001$).

Nevertheless, these results only hint at CALM’s broader potential. We believe that CALM lays the foundation for a new class of systems -- ones that reason with rigor, structure, composability, and interpretability yet adapt with neural perception, bridging the abstract and the real in a formally defined logic.